\documentclass{article}
\usepackage{authblk}
\usepackage{spconf,amsmath,graphicx}
\usepackage{booktabs}
\usepackage{multirow}
\usepackage[table,xcdraw]{xcolor}

\title{ShaDocNet: Learning Spatial-Aware Tokens in Transformer for Document Shadow Removal}
\twoauthors{Xuhang Chen\thanks{This work was supported in part by the University of Macau under Grant MYRG2022-00190-FST, in part by the Science and Technology Development Fund, Macau SAR,
under Grant 0034/2019/AMJ, Grant 0087/2020/A2 and Grant 0049/2021/A, in part by the National Natural Science Foundations of China under Grants 62172403 and in part by the Distinguished Young Scholars Fund of Guangdong under Grant 2021B1515020019.}\sthanks{Equal Contribution} \quad Xiaodong Cun\footnotemark[2] \quad Chi-Man Pun\sthanks{Corresponding author, email: cmpun@umac.mo, sq.wang@siat.ac.cn}}
                           {Department of Computer and Information Science\\University of Macau}
                           {Shuqiang Wang\footnotemark[3]}
                           {Shenzhen  Institute  of  Advanced Technology\\Chinese Academy of Sciences}

\begin{document}
%
\maketitle
\begin{abstract}
Shadow removal improves the visual quality and legibility of digital copies of documents. However, document shadow removal remains an unresolved subject. 
Traditional techniques rely on heuristics that vary from situation to situation. Given the quality and quantity of current public datasets, the majority of neural network models are ill-equipped for this task.
In this paper, we propose a Transformer-based model for document shadow removal that utilizes shadow context encoding and decoding in both shadow and shadow-free regions. Additionally, shadow detection and pixel-level enhancement are included in the whole coarse-to-fine process.
On the basis of comprehensive benchmark evaluations, it is competitive with state-of-the-art methods.
\end{abstract}

\section{Introduction}
\label{sec:intro}

\begin{figure}[ht]
    \begin{minipage}[b]{1.0\linewidth}
        \begin{minipage}[b]{.32\linewidth}
            \centering
            \centerline{\includegraphics[width=\columnwidth,height=3cm]{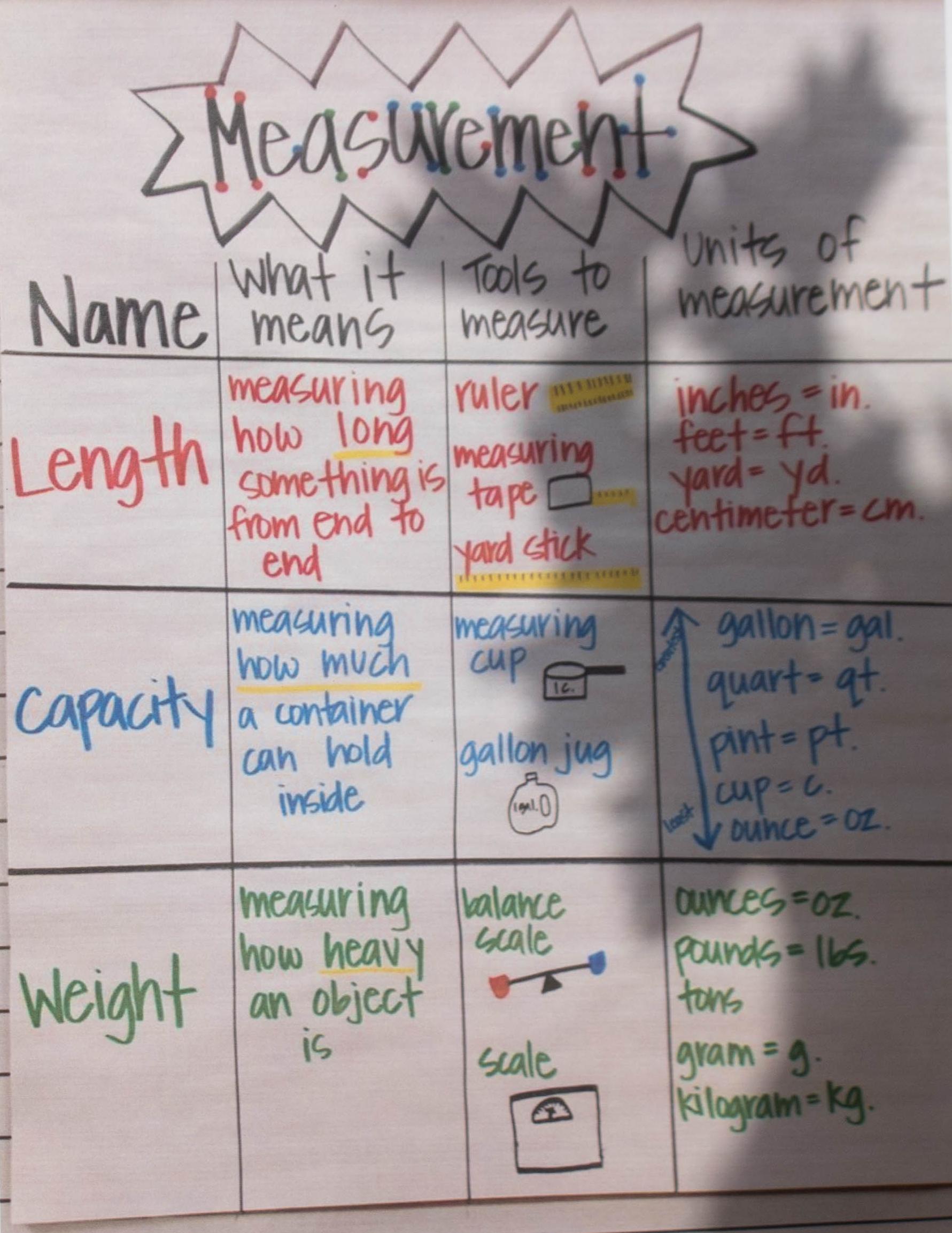}}
            \centerline{(a) Shadow Image}\medskip
        \end{minipage}
        \hfill
        \begin{minipage}[b]{0.32\linewidth}
            \centering
            \centerline{\includegraphics[width=\columnwidth,height=3cm]{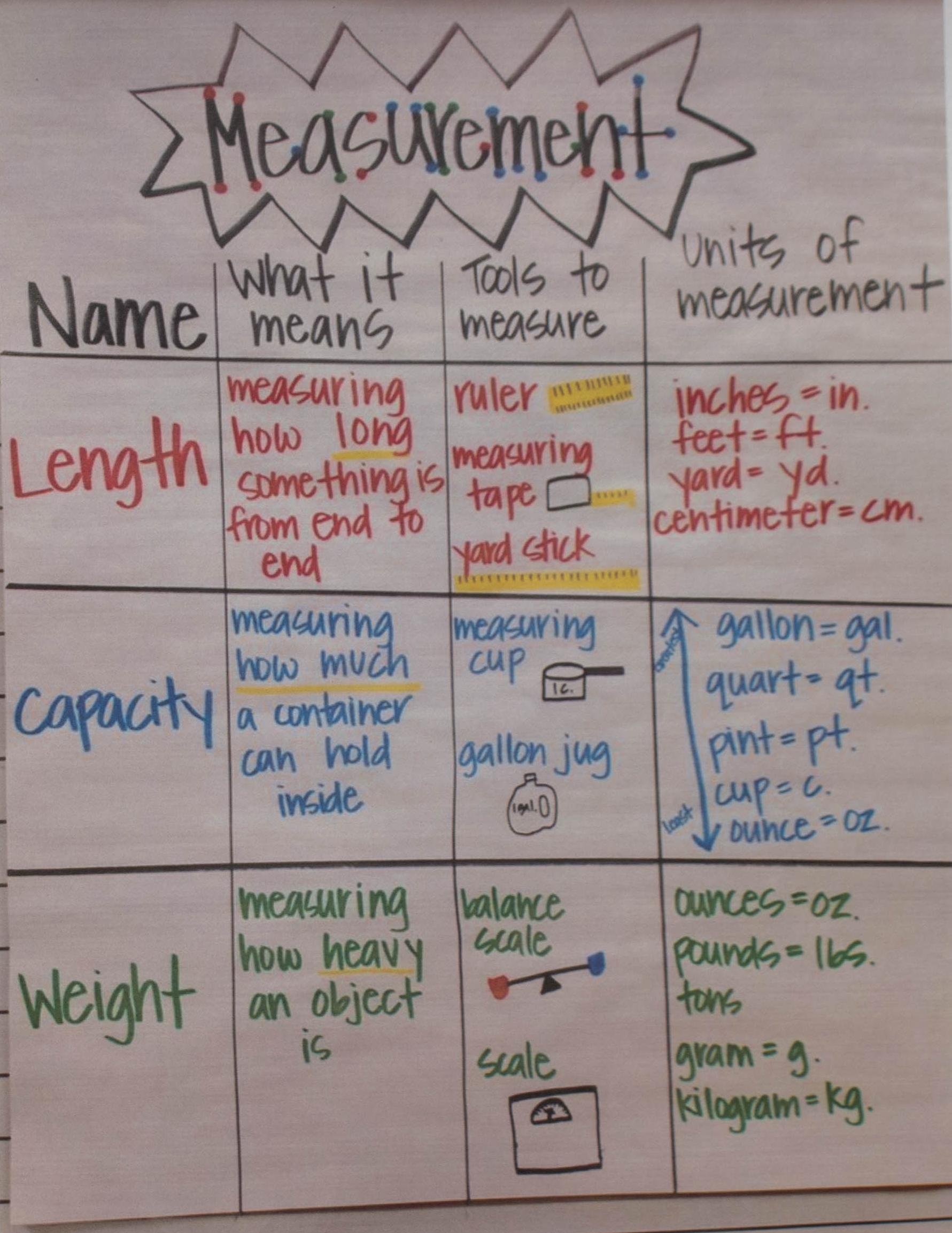}}
            \centerline{(b) Ground Truth}\medskip
        \end{minipage}
        \hfill
        \begin{minipage}[b]{0.32\linewidth}
            \centering
            \centerline{\includegraphics[width=\columnwidth,height=3cm]{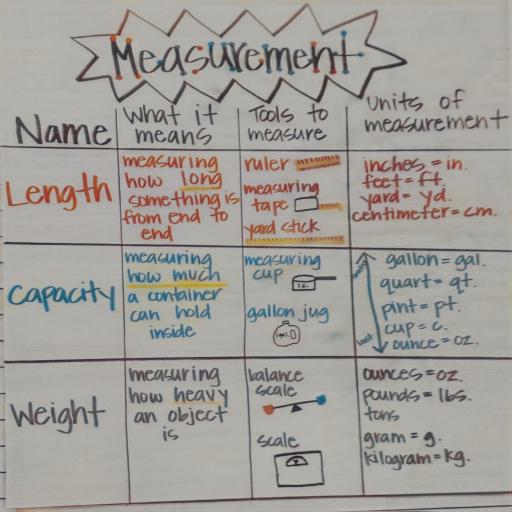}}
            \centerline{(c) ShadocNet}\medskip
        \end{minipage}
    \end{minipage}
    \begin{minipage}[b]{1.0\linewidth}
        \begin{minipage}[b]{.32\linewidth}
            \centering
            \centerline{\includegraphics[width=\columnwidth,height=3cm]{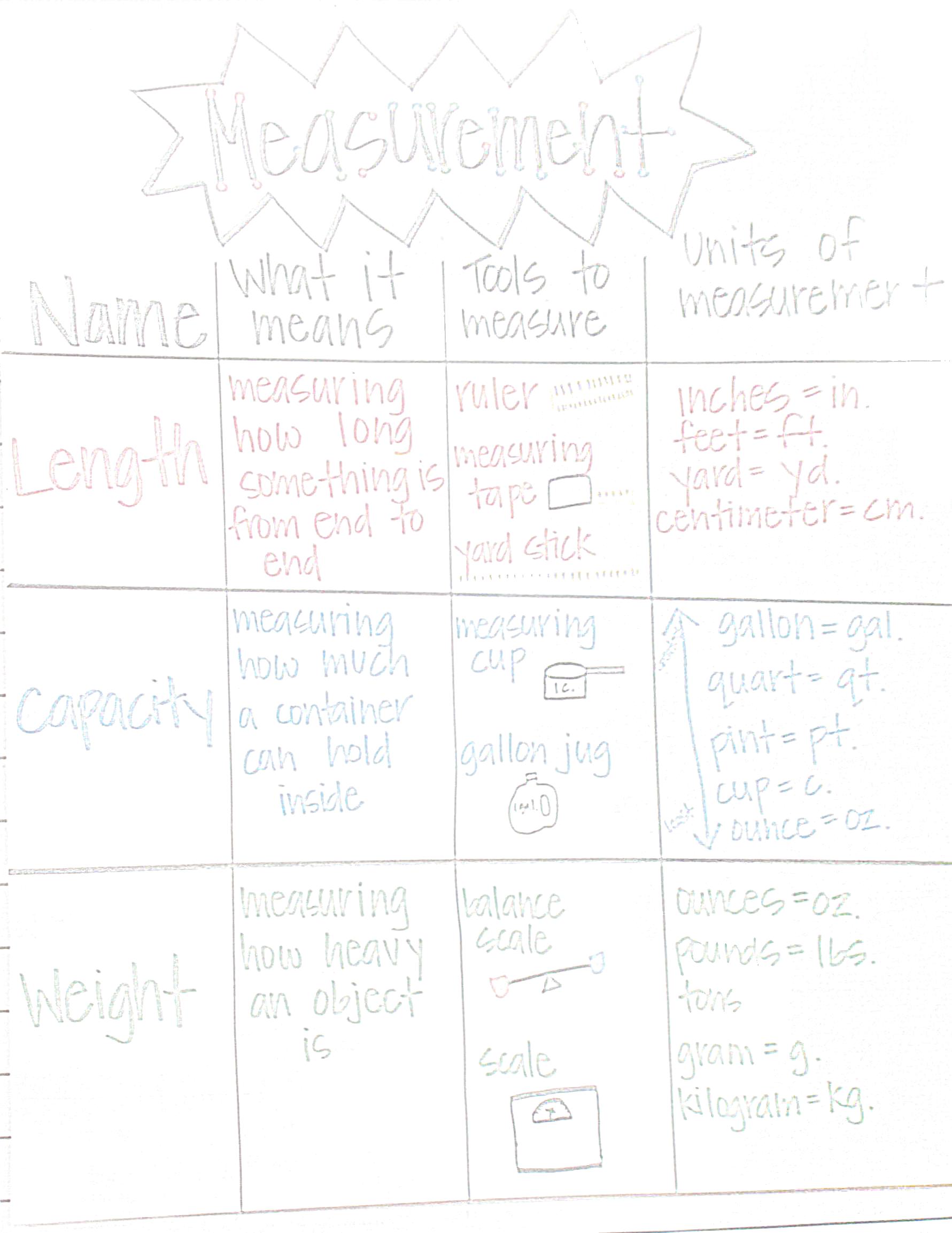}}
            \centerline{(d) Shah \emph{et al.}}\medskip
        \end{minipage}
        \hfill
        \begin{minipage}[b]{0.32\linewidth}
            \centering
            \centerline{\includegraphics[width=\columnwidth,height=3cm]{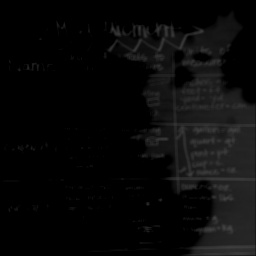}}
            \centerline{(e) AEFNet}\medskip
        \end{minipage}
        \hfill
        \begin{minipage}[b]{0.32\linewidth}
            \centering
            \centerline{\includegraphics[width=\columnwidth,height=3cm]{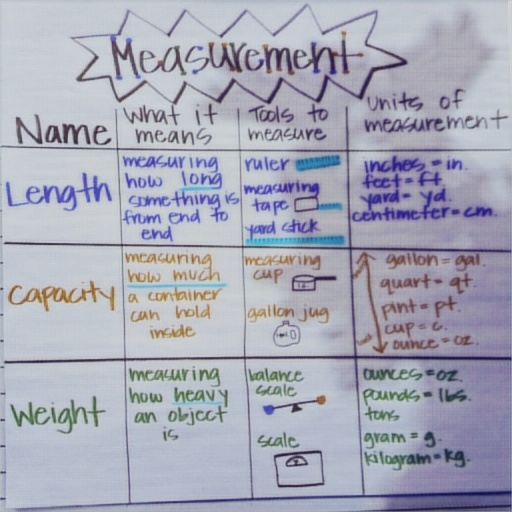}}
            \centerline{(f) BEDSR-Net}\medskip
        \end{minipage}
    \end{minipage}
    \caption{ 
        Previous methods, including traditional method such as Shah \emph{et al.}'s method~\cite{Shah2018AnIA}, or deep learning models, for exmaple, AEFNet~\cite{fu2021auto}, and BEDSR-Net~\cite{lin2020bedsr}, produce results with defects such as shade edges~(f), overexposure~(d) and color fading~(e) in their results. Our model ShadocNet~(c) has much fewer artifacts and is much closer to the ground-truth shadow-free image~(b).
    }
    \label{showcase}
\end{figure}

Documents store and deliver vital information that is necessary in our daily lives. However, casual document photographs acquired in the wild frequently feature shading errors due to different occluders blocking the light sources. Since shadows typically hinder the quality of documents, document shadow removal is an essential computer vision task.

The majority of current document shadow removal methods rely on heuristics to investigate certain features of document images~\cite{jung2018water, kligler2018document}. However, due to the inherent limits of artificial heuristics, they often perform well for certain images but not others. As a consequence, their outputs often display different forms of defects for document images as shown in Figure~\ref{showcase}~(d). 

Recent years have witnessed the introduction of a few document shadow removal methods based on deep learning. By using a large synthetic dataset, BEDSR-Net mitigates shadow based on the prediction of the background color and shadow attention map~\cite{lin2020bedsr}. It is the first deep network created exclusively for removing document image shadows. However, BEDSR-Net is unable to completely recover a proper shadow-free image, as shown in Figure~\ref{showcase}~(f).

\begin{figure*}[ht]
    \centering
    \includegraphics[width=\textwidth]{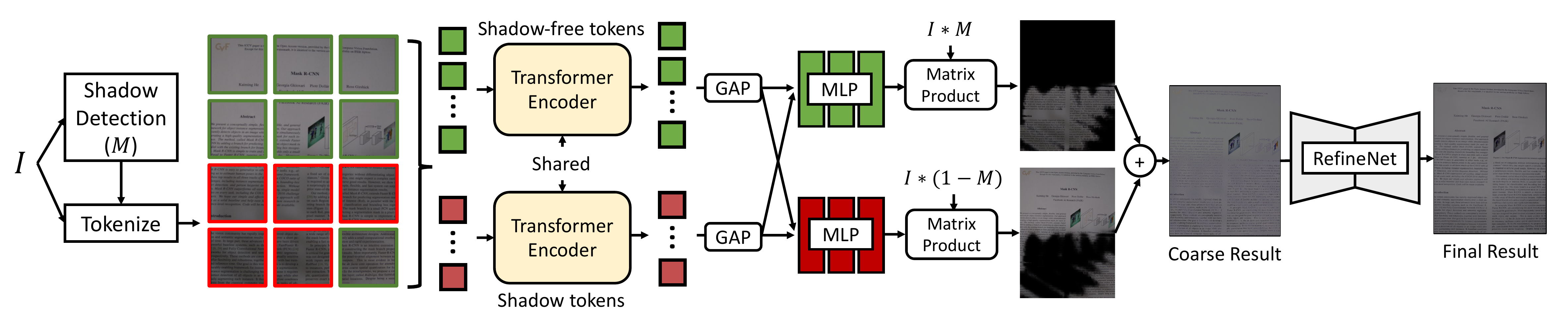}
    \caption{The network structure of the ShadocNet. It consists of three components: the Shadow Detection module, the Global Shadow Remapping for shadow color remapping, and the RefineNet for improving visual quality.}
    \label{fig:overall}
\end{figure*}

Intuitively, shadow removal methods for natural images seems can be transferable for document images. In reality, there are two issues with using aforementioned models to remove document shadows. For starters, they mostly demand a large dataset to train with. Second, since these approaches do not utilize specific properties of document images, performance would be inferior in this domain shift circumstance. As illustrated in Figure~\ref{showcase}~(e), AEFNet is not capable to recover a shadow-free document even after training with document shadow images and the corresponding shadow masks.

This paper proposes a new Transformer-based model ShadocNet for document shadow removal. Extensive benchmarks demonstrates that our model beats previous methods in terms of aesthetic quality, as shown in Figure~\ref{showcase}~(c). We summarize our major contributions as follows.

\begin{enumerate}
\item We propose a robust model ShadocNet for document shadow removal, which outperforms state-of-the-art methods. 
\item We are the pioneers of document shadow elimination using transformer architecture.
\end{enumerate}

\section{Related Work}
\label{sec:related}

Certain techniques have been developed expressly for the purpose of removing shadows from document photographs. Jung proposed both the water-filling method and the document shadow dataset~\cite{jung2018water}. Kligler provided the other dataset along with a deshadow method inspired by a 3D point cloud~\cite{kligler2018document}. Other traditional methods make use of the estimation of document background color~\cite{wang2019effective, Wang2020ShadowRO} or shading and reflectance effects~\cite{Shah2018AnIA}. Nevertheless, shadow residues are often left in their outputs.

BEDSR-Net is the state-of-the-art model in document shadow removal~\cite{lin2020bedsr}. BEDSR-Net includes a Background Estimation Network (BE-Net) for predicting the document's global background color and a shadow removal module that operates shadow removal significantly better using both the predicted background color and the attention map.

Methodologies for depicting nature shadow images may be informative and transferable in document images. ST-CGAN applies two stacked GAN~\cite{you2022fine} to remove shadow~\cite{wang2018stacked}. By estimating the over-exposure image condition and combining the original input with the over-exposure shadow region, AEFNet align the color of shadow areas with that of shadow-free parts~\cite{fu2021auto}. Mask-ShadowNet is built on a well-designed masked adaptive instance normalization technique with incorporated aligners~\cite{he2021mask}.

\section{Proposed Method}
\label{sec:method}

We propose the ShadocNet network to assess the document shadow removal method on ShaDocs. As shown in Figure~\ref{fig:overall}, we build a multi-stage framework for document shadow removal that includes shadow detection, global color matching, and local pixel-by-pixel refining. Listed below are the specifics of each component and loss function.

\subsection{The structure of ShadocNet}
\subsubsection{Shadow detection}
Since most of the previous shadow detection methods are designed for natural shadow removal, we retrain a learning-based method~\cite{Zheng2019DistractionAwareSD} to extract the shadow masks on the document images. It starts with extracting hierarchical feature maps that encode fine details and semantic information. Then a direction-aware spatial context (DSC) module gathers directed spatial contexts. The DSC and convolutional features are concatenated and upsampled to original image size. It integrates upsampled feature maps into multi-level integrated features using a convolution layer and deep supervision to predict a score map at each layer. The final shadow map incorporates all predicted score maps.

\begin{table*}[t]
    \centering
    \resizebox{\textwidth}{!}
    {
        \begin{tabular}{c|c|cccccccccccc}
            \hline
                                                &                           & \multicolumn{12}{c}{Methods}                                                                                                                                                                                                                                                                                                                                                                   \\ \cline{3-14}
            \multirow{-2}{*}{Dataset}           & \multirow{-2}{*}{Metrics} & Jung \cite{jung2018water}            & Shah \cite{Shah2018AnIA} & Wang \cite{wang2019effective} & Wang \cite{Wang2020ShadowRO} & U-Net \cite{ronneberger2015u} & ST-CGAN \cite{wang2018stacked} & BEDSR-Net \cite{lin2020bedsr} & AEFNet \cite{fu2021auto} & \multicolumn{1}{c|}{Mask-ShadowNet \cite{he2021mask}} & Ours w/o ViT & Ours w/o RefineNet & Ours                                  \\ \hline
                                                & RMSE$\downarrow$          & 17.88                                & 46.77                    & 40.50                         & 72.78                        & 20.44                         & 86.52                          & 23.36                         & 208.41                   & \multicolumn{1}{c|}{45.87}                            & 18.49        & 27.04              & {\color[HTML]{FE0000} \textbf{15.30}} \\
                                                & PSNR$\uparrow$            & 23.27                                & 14.91                    & 16.86                         & 11.23                        & 22.47                         & 9.40                           & 21.98                         & 1.76                     & \multicolumn{1}{c|}{15.51}                            & 23.00        & 19.86              & {\color[HTML]{FE0000} \textbf{24.60}} \\
            \multirow{-3}{*}{Jung's dataset}    & SSIM$\uparrow$            & {\color[HTML]{FE0000} \textbf{0.91}} & 0.83                     & 0.86                          & 0.81                         & 0.86                          & 0.35                           & 0.88                          & 0.00                     & \multicolumn{1}{c|}{0.78}                             & 0.86         & 0.78               & {\color[HTML]{FE0000} \textbf{0.91}}  \\ \hline
                                                & RMSE$\downarrow$          & 49.15                                & 97.48                    & 19.35                         & 43.43                        & 21.58                         & 37.29                          & 18.78                         & 146.06                   & \multicolumn{1}{c|}{24.92}                            & 14.47        & 23.51              & {\color[HTML]{FE0000} \textbf{13.48}} \\
                                                & PSNR$\uparrow$            & 14.44                                & 8.39                     & 22.97                         & 15.93                        & 21.68                         & 16.85                          & 24.37                         & 4.86                     & \multicolumn{1}{c|}{20.48}                            & 25.98        & 20.97              & {\color[HTML]{FE0000} \textbf{26.20}} \\
            \multirow{-3}{*}{Kligler's dataset} & SSIM$\uparrow$            & 0.90                                 & 0.70                     & 0.88                          & 0.85                         & 0.83                          & 0.54                           & 0.88                          & 0.05                     & \multicolumn{1}{c|}{0.81}                             & 0.92         & 0.71               & {\color[HTML]{FE0000} \textbf{0.94}}  \\ \hline
        \end{tabular}
    }
    \caption{Quantitative comparisons of visual quality using RMSE, PSNR and SSIM. We compare our model with nine competitive methods and two variants of ShadocNet. The top scores are highlighted in red and bold.}
    \label{quan}
\end{table*}

\subsubsection{Global shadow remapping}
After single image shadow extraction, we aim to remap the color of shadowed region based on the assumption that the cast shadow is formed by a uniform illuminant, inspired by image harmonization~cite{liang2021spatial}. Thus, we apply the vision transformer~(ViT)~\cite{dosovitskiy2020image} to accomplish the aforementioned objective. Specifically, we tokenize the original images to the patch-based tokens firstly, and then, we utilize the extracted mask from the previous step to divide all of the patch embeddings into two groups, which contains only the shadow or the shadow-free token embeddings. 

Following this, we compute the self-attention on these domain-aware patch embeddings using the transformer, resulting in a domain-aware region embedding for the shadow and shadow-free area. We employ a global average pooling layer~(GAP) to construct the foreground and background global representations, since the mask still includes error segmentation and the patch-based representation is not pixel-wise.

After obtaining the domain-aware tokens, we develop the MLP, a simple yet effective pixel mapping function to remap each region's original color pixels to their new value. Since shadows often result from homogeneous illumination, the color variations in the shadow area will be comparable. Therefore, MLP transfers the per-pixel values to the restored shadow-free region in order to recreate the image.

\subsubsection{RefineNet} 

Inspired by Dual Hierarchical Aggregation Network~\cite{Cun2020TowardsGS}, we also apply a multi-stage convolution and aggregation strategy for pixel-wise refinement. In detail, we first use a pre-trained VGG16 to extract hyper-column feature. Then, to derive spatial attentions and mixed layer features, in each level of aggregation, we apply a squeeze-and-excitation block~\cite{hu2018squeeze} to re-weight feature channels. Finally, a spatial pooling pyramid is added at the conclusion of the last aggregation block for multi-context features remixing.

\subsection{Loss function}
We consider document shadow removal as a supervised issue. In particular, we calculate the difference between the target and the corresponding produced images~(for each step) in the composited area. Consequently, we use relative $L_{1}$ loss between the predicted foreground and the target with respect to the foreground shadow mask $M$. In addition, for improved visual quality, we include multi-layer perception loss~\cite{zhang2018single} into our framework. Here are the specifics of each component.

\subsubsection{Relative $L_{1}$ loss $L_{pixel}$}

We compute the metric between the predicted image's foreground and the target, where differences are only quantified inside a single domain. Inspiring by previous studies in watermark removal~\cite{hertz2019blind}, we thus perform pixel-by-pixel $L_{1}$ loss in both the foreground shadow masked area $M$ and the background document unmasked area $M' = 1 - M$ by masking away the counterpart pixels and establishing the relevant region. Particularly, providing the generated images $I_{n}$ in two stages, global shadow remapping and RefineNet, the loss across the masked area is computed.

\begin{equation}
\begin{aligned}
\label{l1_loss} 
L_{pixel}=\sum_{n=1}^{N} \frac{\left ||M \times I_{n}-M \times I_{gt}\right ||_{1}}{sum(M)} \\
+ \sum_{n=1}^{N} \frac{\left ||M' \times I_{n}-M' \times I_{gt}\right ||_{1}}{sum(M')}
\end{aligned}
\end{equation}

where $N=2$ is the number of stages.

\subsubsection{Perception loss $L_{\Phi}$}

By considering the semantic measures and low-level details in multiple contexts, we also introduce multi layer perception loss $L_{\Phi}$ with a weight $\alpha$. The formula of perception loss is as follows:

\begin{equation}\label{perc_loss} 
  L_{\Phi} = \sum^{5}_{k=0}\lambda_{l}||\Phi_{k}( I_{free}^{'})) - \Phi_{k}(I_{free}) ||_1,
\end{equation}

where $\Phi$ is the VGG16 network pre-trained on ImageNet.

Overall, our model can be trained in an objective function by combining the aforementioned losses:

\begin{equation}\label{total_loss}
    L_{total} = \lambda_{pixel} L_{pixel} + \lambda_{\Phi} L_{\Phi}
\end{equation}

\begin{figure*}[t]
\begin{minipage}[b]{1.0\linewidth}
    \begin{minipage}[b]{0.12\linewidth}
        \centering
        \centerline{\includegraphics[width=\columnwidth,height=4.8cm]{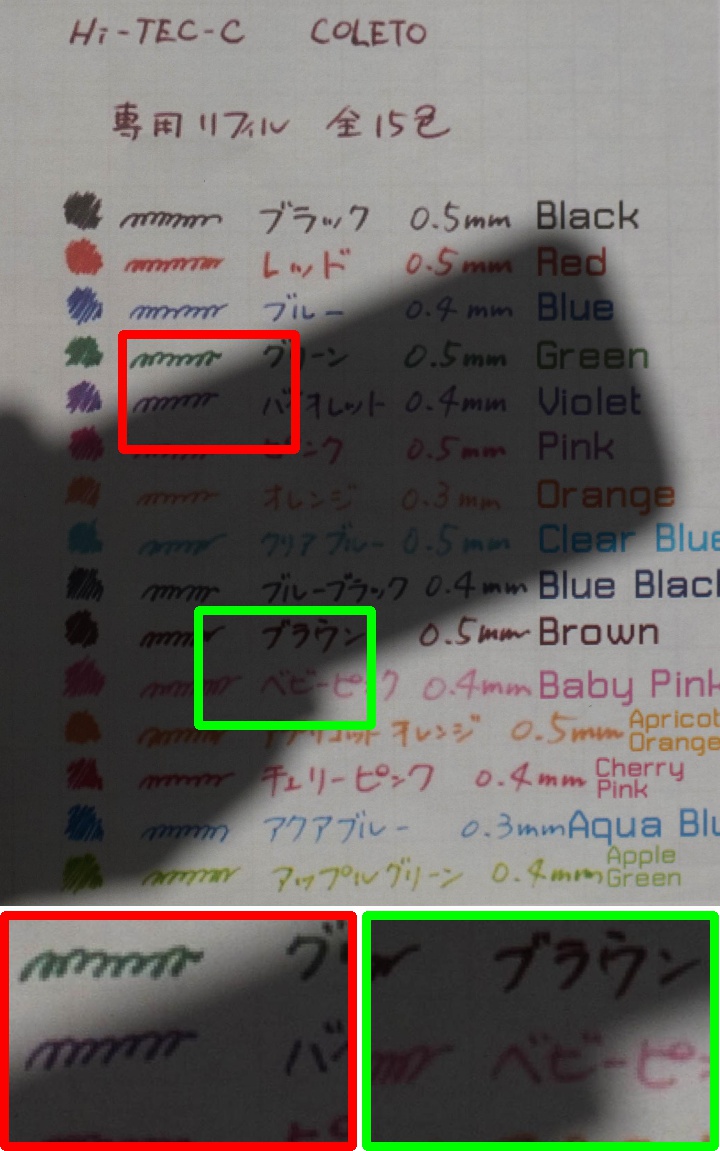}}
        \centerline{(a)}\medskip
    \end{minipage}
    \hfill
    \begin{minipage}[b]{0.12\linewidth}
        \centering
        \centerline{\includegraphics[width=\columnwidth,height=4.8cm]{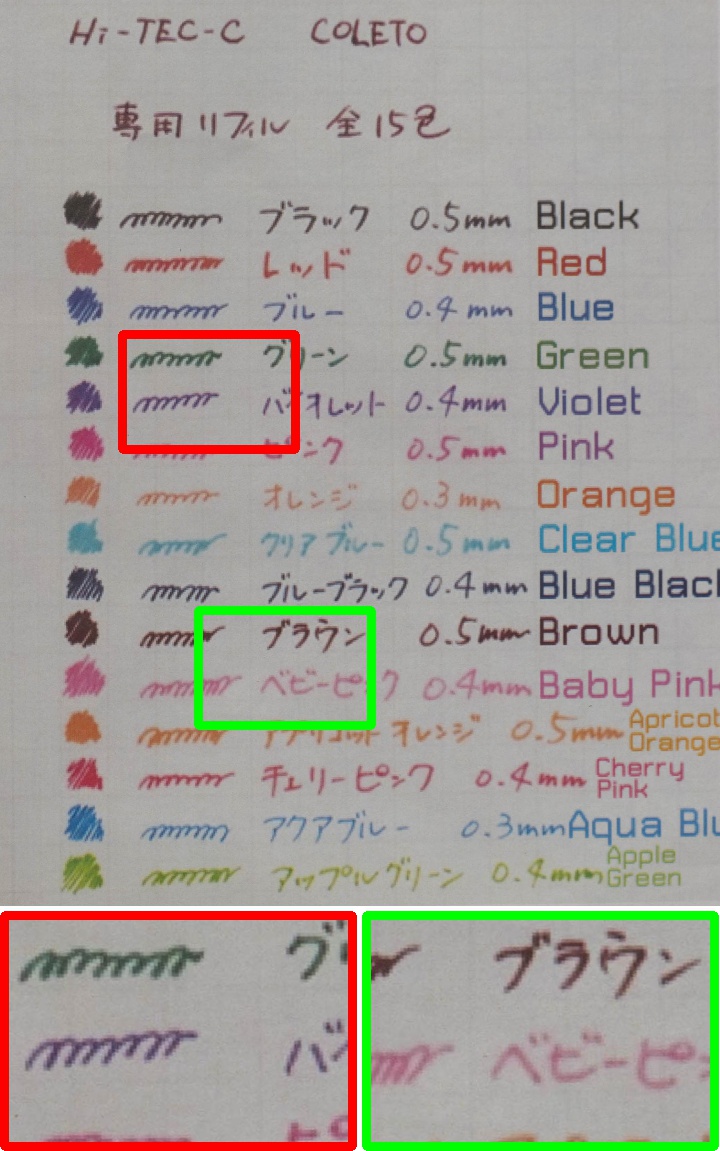}}
        \centerline{(b)}\medskip
    \end{minipage}
    \hfill
    \begin{minipage}[b]{0.12\linewidth}
        \centering
        \centerline{\includegraphics[width=\columnwidth,height=4.8cm]{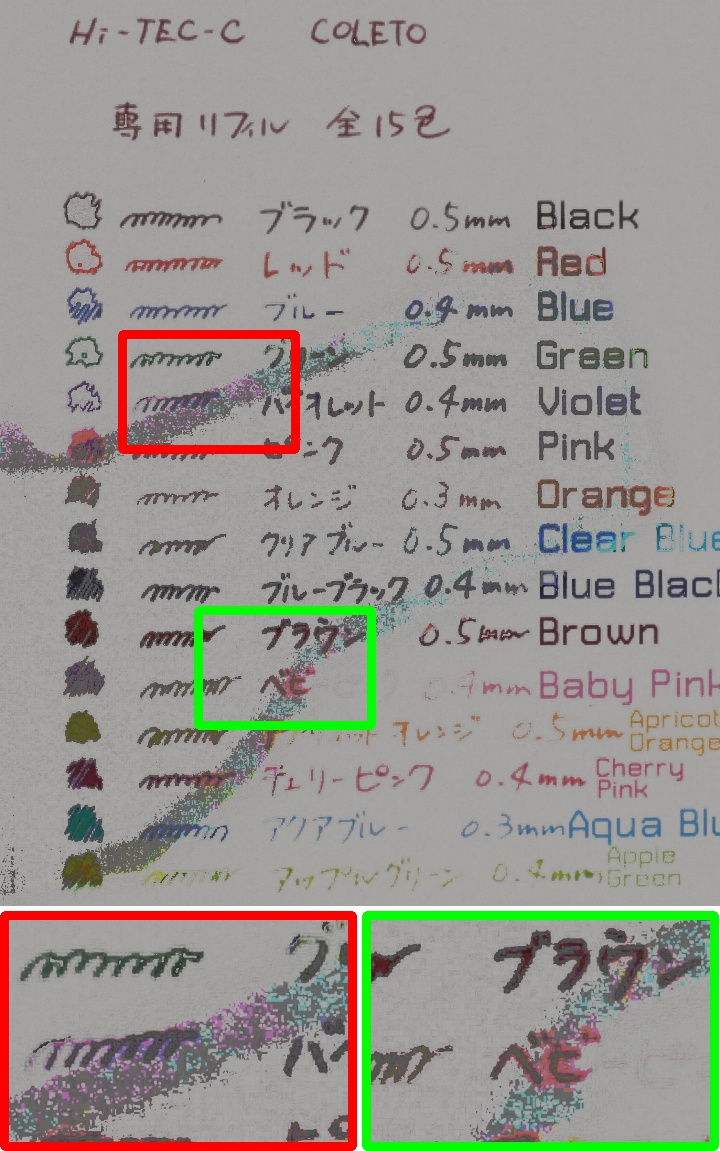}}
        \centerline{(c)}\medskip
    \end{minipage}
    \hfill
    \begin{minipage}[b]{0.12\linewidth}
        \centering
        \centerline{\includegraphics[width=\columnwidth,height=4.8cm]{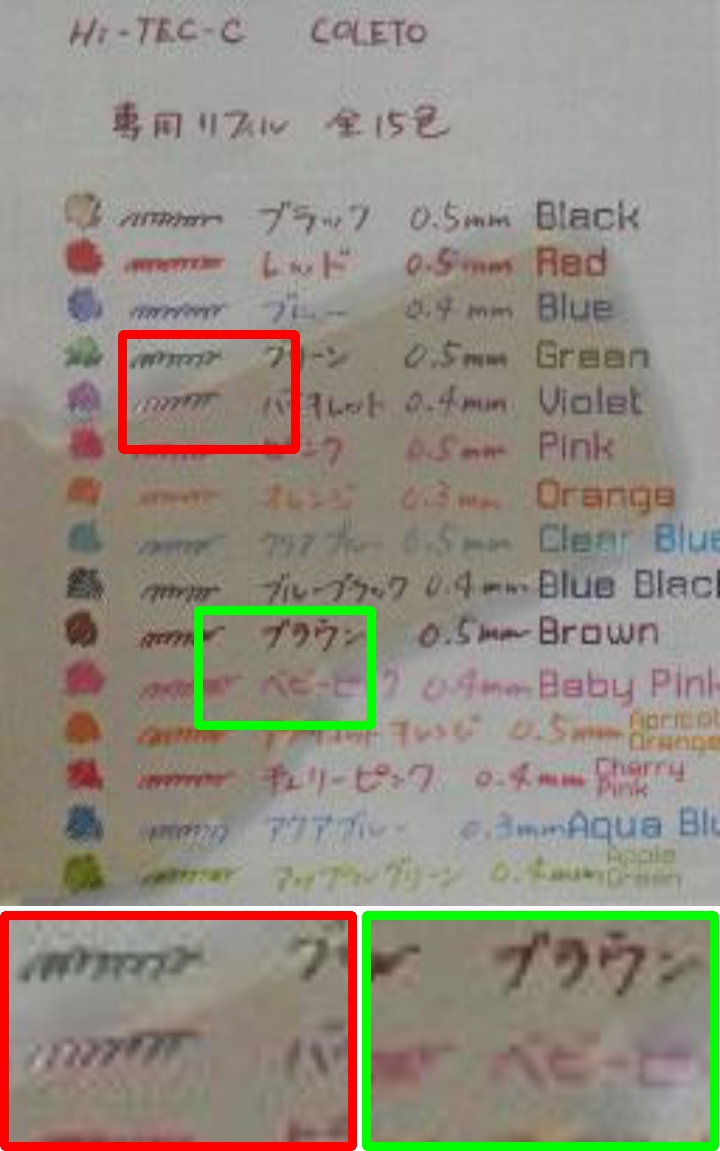}}
        \centerline{(d)}\medskip
    \end{minipage}
    \hfill
    \begin{minipage}[b]{0.12\linewidth}
        \centering
        \centerline{\includegraphics[width=\columnwidth,height=4.8cm]{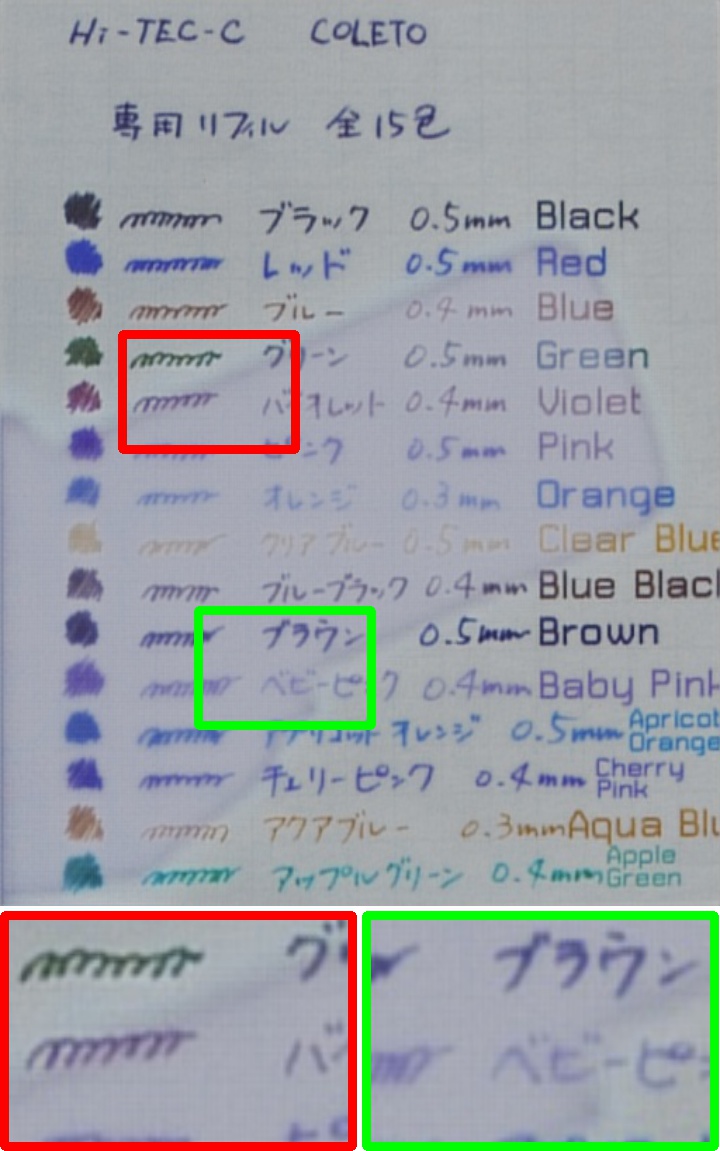}}
        \centerline{(e)}\medskip
    \end{minipage}
    \hfill
    \begin{minipage}[b]{0.12\linewidth}
        \centering
        \centerline{\includegraphics[width=\columnwidth,height=4.8cm]{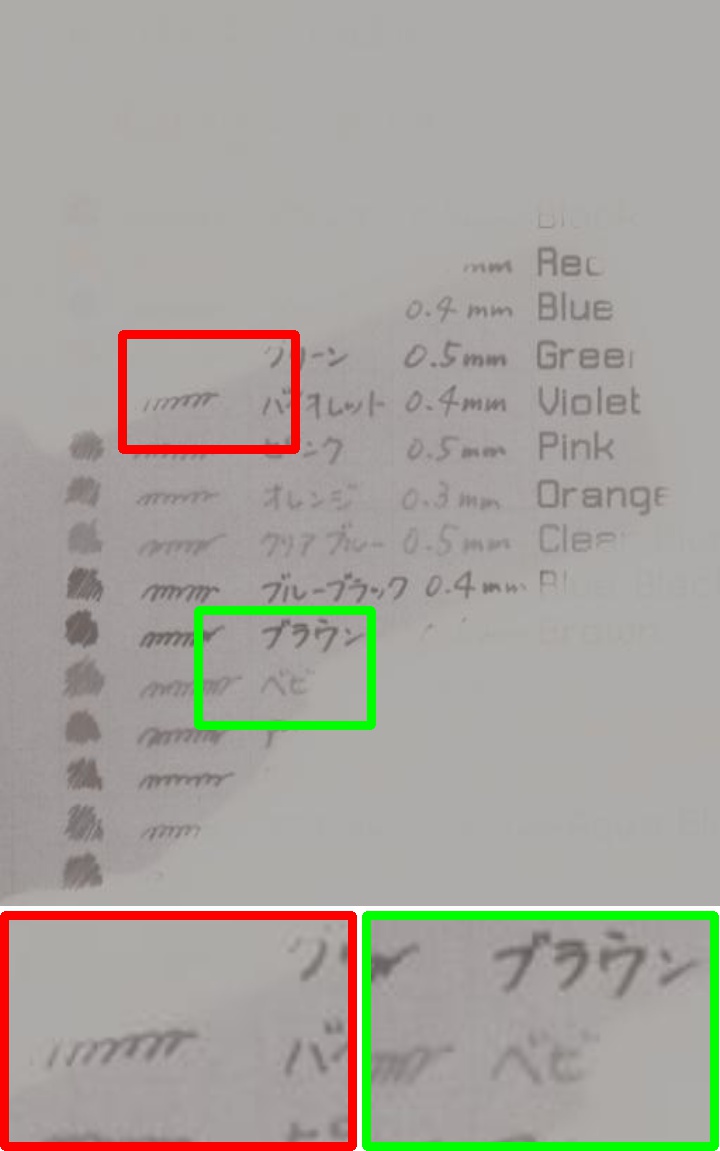}}
        \centerline{(f)}\medskip
    \end{minipage}
    \hfill
    \begin{minipage}[b]{0.12\linewidth}
        \centering
        \centerline{\includegraphics[width=\columnwidth,height=4.8cm]{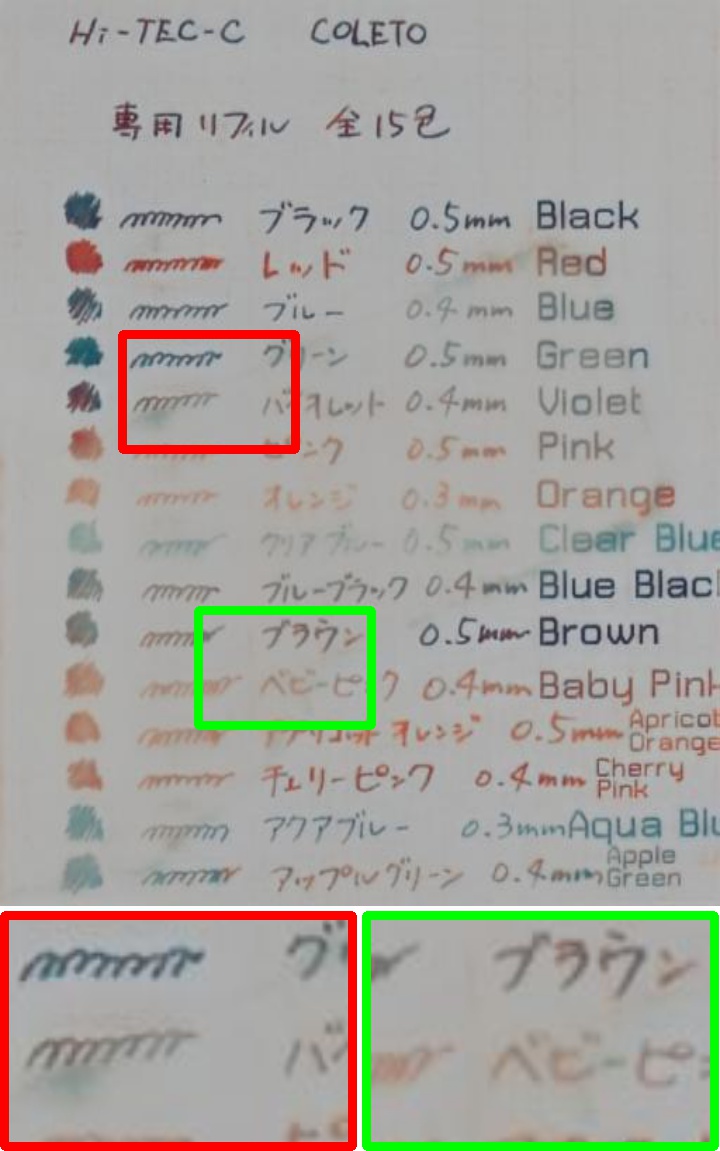}}
        \centerline{(g)}\medskip
    \end{minipage}
    \hfill
    \begin{minipage}[b]{0.12\linewidth}
        \centering
        \centerline{\includegraphics[width=\columnwidth,height=4.8cm]{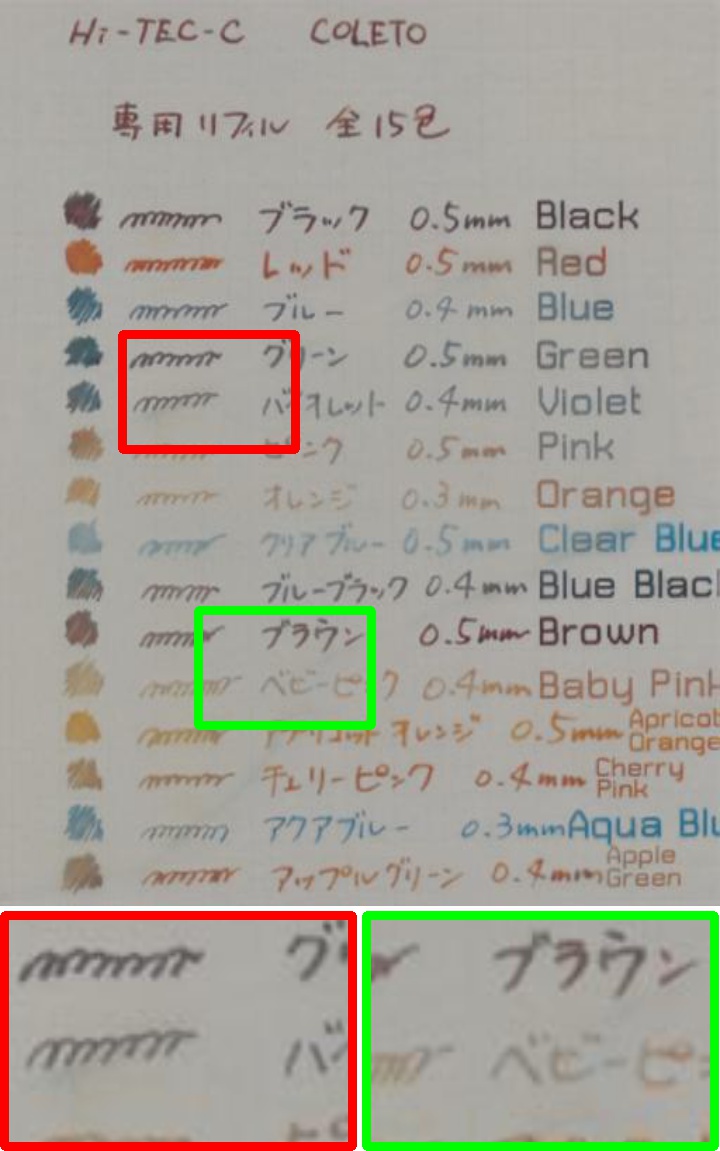}}
        \centerline{(h)}\medskip
    \end{minipage}
\end{minipage}
\caption{Visual comparison of competing methods. From left to right:~(a) shadow image,~(b) ground truth,~(c) results of Wang~\emph{et al.}~\cite{wang2019effective},~(d) results of Mask-ShadowNet \cite{he2021mask},~(e) results of BEDSR-Net~\cite{lin2020bedsr},~(f) results of ShadocNet w/o RefineNet,~(g) results of ShadocNet w/o Transformer encoder,~(h) results of ShadocNet.}
\label{visual-compare}
\end{figure*}

\section{Experiments}
\label{sec:exp}

We compare different document shadow removal methods and quantitative metrics based on visual quality and content recovery.

\subsection{Comparisons with state-of-the-art methods}

Our model is compared to seven state-of-the-art approaches, including traditional document shadow removal methods~\cite{jung2018water, Shah2018AnIA, wang2019effective, Wang2020ShadowRO} and several state-of-the-art deep learning-based shadow removal methods including ST-CGAN~\cite{wang2018stacked}, BEDSR-Net~\cite{lin2020bedsr}, Mask-ShadowNet~\cite{he2021mask} and AEFNet~\cite{fu2021auto}. Additionally, we use U-Net~\cite{ronneberger2015u,hu2019cross} performance as a benchmark for improved demonstration. We utilized the authors' publicly accessible source codes wherever they were available for a fair comparison.

We employ the Root Mean Square Error~(RMSE), Peak Signal-to-Noise Ratio~(PSNR), and Structural Similarity~(SSIM) measures to evaluate visual quality. RMSE is the most important and widely used metric in shadow removal, while PSNR and SSIM are frequently used in image restoration evaluations and low-level computer vision tasks. For content preservation assessment, we evaluate the effectiveness of Optical Character Recognition~(OCR) techniques on recovered images devoid of shadows. In principle, OCR should be able to identify more content if the document is better recovered.

\subsection{Quantitative evaluation}

Table~\ref{quan} provides quantitative comparisons of the compared algorithms on two datasets, as well as the average RMSE, PSNR, and SSIM values.

Traditional methods depend heavily on heuristics and cannot perform effectively over a variety of datasets. 

Due to the fact that certain models require a large training dataset and only capture features in natural images, they are incapable of removing document shadows.

Compared to the aforementioned methods, our model can extract shadow area tokens and shadow-free region tokens with the assistance of our detection module and ViT encoder. Using a two-stage coarse-to-fine approach, ShadocNet is more resilient than those of the previous since it yields steady and good results for images with different characteristics.

Our model derives from ViT. As an ablation study, table~\ref{quan} presented the performance of our model without ViT decreases, demonstrating that our performance is not only dependent on the RefineNet design. As an additional ablation study, the superior benchmark of our model without ViT shows the effectiveness of the RefineNet. 

\subsection{Evaluation on content preservation}

In addition to quantitative metrics, we report the OCR performance on the recovered shadow-free images. First, we use an open-source OCR program~\cite{jaided2022jaided} to detect words for shadow-free ground-truth images and the outcomes of comparing techniques. Then, the OCR performance is evaluated by comparing the text strings using the Levenshtein distance, also known as edit-distance.

As shown in Table~\ref{ocr} and Figure~\ref{visual-compare}, ShadocNet surpasses its competitors, demonstrating that it also improves the readability of texts by retaining their content more accurately.

\begin{table}[ht]
\centering
{\begin{tabular}{l|l}
\hline
method & edit distance$\downarrow$ \\ \hline

Wang \emph{et al.} \cite{wang2019effective} & 191.5 \\

Shah \emph{et al.} \cite{Shah2018AnIA} & 187.8 \\

U-Net \cite{ronneberger2015u} & 261.2 \\

ST-CGAN \cite{wang2018stacked} & 294.8 \\

Mask-ShadowNet \cite{he2021mask} & 262.9 \\

AEFNet \cite{fu2021auto} & 292.4 \\

BEDSR-Net \cite{lin2020bedsr} & 189.0 \\

\hline

ShadocNet w/o Vit & 191.6 \\

ShadocNet w/o RefineNet & 244.2 \\

ShadocNet & \textbf{178.6}

\\\hline

\end{tabular}}
\caption{
Average OCR edit-distances of ShadocNet and its two variants, Wang \emph{et al.}'s method, Shah \emph{et al.}'s method, U-Net, ST-CGAN, Mask-ShadowNet, AEFNet and BEDSR-Net.
}
\label{ocr}
\end{table}

\section{Conclusion}
\label{sec:conclusion}

In this paper, we propose a competitive Transformer-based ShadocNet by leveraging the feature extractor and color rendering module to boost the visual quality. Extensive experiments indicate that the proposed ShadocNet performs favorably against state-of-the-art methods on various dataset. In the future, we are going to maintain a continued evolution of ShadocNet, including the scale, quality and diversity.

\vfill\pagebreak
\bibliographystyle{IEEEbib}
\bibliography{refs}

\end{document}